\documentclass{article}

\usepackage{PRIMEarxiv}

\usepackage[utf8]{inputenc} 
\usepackage[T1]{fontenc}    
\usepackage{hyperref}       
\usepackage{url}            
\usepackage{booktabs}       
\usepackage{amsfonts}       
\usepackage{nicefrac}       
\usepackage{microtype}      
\usepackage{lipsum}
\usepackage{fancyhdr}       
\usepackage{graphicx}       
\graphicspath{{media/}}     
\usepackage{amsmath}
\usepackage{gensymb}
\usepackage{longtable}
\usepackage{subcaption}
\usepackage{pdfpages}
\usepackage{longtable}
\usepackage{setspace}
\usepackage{color}
\usepackage{cite} 

\title{Domain Consistent Industrial Decarbonisation of Global Coal Power Plants
}

\author{
  Waqar Muhammad Ashraf* \\
  University College London \\
  London, UK\\
  University of Cambridge \\
  Cambridge, UK\\
  and \\
  The Alan Turing Institute \\
  London, UK\\
  \texttt{*waqar.ashraf.21@ucl.ac.uk}\\
    \AND
  Vivek Dua \\
  University College London \\
  London, UK\\
  \texttt{v.dua@ucl.ac.uk}\\
  \And
  Ramit Debnath* \\
  University of Cambridge \\
  Cambridge, UK\\ 
  \texttt{*rd545@cam.ac.uk} \\
}

\begin{document}
\maketitle

\begin{abstract}
Machine learning and optimisation techniques (MLOPT) hold significant potential to accelerate the decarbonisation of industrial systems by enabling data-driven operational improvements. However, the practical application of MLOPT in industrial settings is often hindered by a lack of domain compliance and system-specific consistency, resulting in suboptimal solutions with limited real-world applicability. To address this challenge, we propose a novel human-in-the-loop (HITL) constraint-based optimisation framework that integrates domain expertise with data-driven methods, ensuring solutions are both technically sound and operationally feasible. We demonstrate the efficacy of this framework through a case study focused on enhancing the thermal efficiency and reducing the turbine heat rate of a 660 MW supercritical coal-fired power plant. By embedding domain knowledge as constraints within the optimisation process, our approach yields solutions that align with the plant's operational patterns and are seamlessly integrated into its control systems. Empirical validation confirms a mean improvement in thermal efficiency of 0.64\% and a mean reduction in turbine heat rate of 93 kJ/kWh. Scaling our analysis to 59 global coal power plants with comparable capacity and fuel type, we estimate a cumulative lifetime reduction of 156.4 million tons of carbon emissions. These results underscore the transformative potential of our HITL-MLOPT framework in delivering domain-compliant, implementable solutions for industrial decarbonisation, offering a scalable pathway to mitigate the environmental impact of coal-based power generation worldwide.
\end{abstract}


\section{Introduction}

Fossil fuel-based thermal plants operating on coal, oil, and gas are the major drivers of greenhouse gas emissions from the energy sector \cite{guo2024carbon}\cite{le2019drivers}. In 2022, fossil fuels generated 81\% of the total energy supply, while coal, oil, and gas contributed 45\%, 33\% and 22\% of global emissions from fuel combustion, respectively \cite{iea}. Despite governments' pledges to phase out or reduce the reliance on fossil fuels for power generation \cite{nacke2024compensating}\cite{jewell2019prospects} and the unreliable energy supply from intermittent renewable power sources \cite{heylen2018review}, achieving net zero emissions requires addressing techno-economic, socio-political, and behavioural challenges around the world \cite{zhang2024targeting}\cite{fankhauser2022meaning}. Although technology upgrade, co-combustion, carbon capture and storage, and early retirement are potential solutions for low-carbon power generation from coal power plants; however, these solutions are expensive and can exert an additional financial burden on low-income and developing countries where fossil-based power plants are being installed to meet the energy demand of increasing populations \cite{foster2024development}\cite{peters2024sustainable}. The early retirement of coal-based energy assets is socially unjust \cite{cui2019quantifying} as most coal-based power plants in Europe and the US are nearing the end of their useful life and have an average life of 35 years and 40 years, respectively \cite{ieacleanenergy}. On the other hand, the average life of coal-based power plants in China is 13 years, and 16 years in the rest of Asian countries, which house a large proportion of coal-based power plants \cite{ieacleanenergy}.

In the spectrum of geopolitical, economic and social challenges in phasing out or low-carbon power generation from coal plants, the International Energy Agency (IEA) proposes an immediate and implementable solution to support industrial decarbonisation by cutting emissions volumes through operational excellence and informed operating strategies built on efficient and reliable analytics for the operation of power generation of fossil energy assets \cite{ieaenergy}\cite{ieaML}. As a policy instrument, the IEA recommendation to improve the operational efficiency of coal power plants appears to be an implementable solution to improve energy efficiency in line with the reduction of CO$_2$ emissions for industrial decarbonisation \cite{yang2025rapid} and appears to be a sustainable approach considering the progress in hardware computing and data-driven analysis techniques. 

\begin{figure}[htp]
    \centering
    \includegraphics[width=1.0\linewidth]{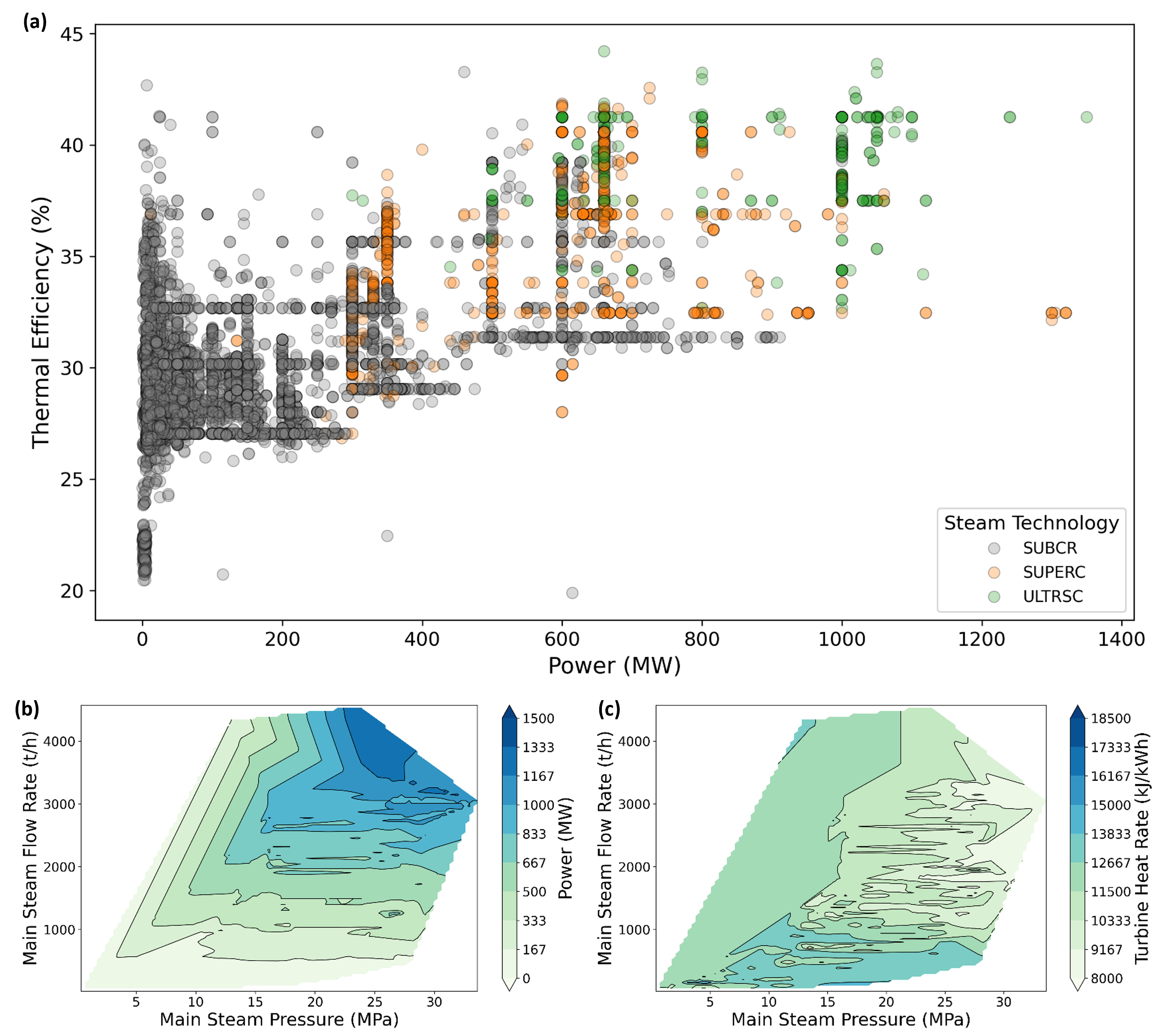}
    \caption{Design parameters of global coal power plants. (a) Major coal-based power plants operate on sub-critical technology with a median energy efficiency of 29.8\% \cite{Sp}\cite{Globalcoal}. Main steam pressure and main steam flow rate are the two important operating parameters and the contour plots plotted against two performance parameters (b) Power (MW) and (c) Turbine Heat Rate (kJ/kWh) represent the highly nonlinear function space for optimising the power generation operation.}
    \label{Fig:global coal}
\end{figure}

However, to demonstrate the complexity bound with the operational excellence of multicapacity and multitechnology coal power plants at the global scale, we have compiled a dataset taken from \cite{Sp}\cite{Globalcoal}, which contains observations on the design specifications of nearly 8870 global coal power plants. The dataset reveals that 80.2\% of the global coal power plants operate on subcritical technology (SUBCR) with a median power generation capacity of 110 megaWatt (MW), as shown in Fig.~\ref{Fig:global coal}(a). In addition, the global median energy efficiency is found to be 29.8\%, 36.9\%, and 40.1\% for SUBCR, supercritical (SUPERC) and ultrasupercritical (ULTRSC) coal power plants. SUBCR coal power plants dominate the global coal-based energy assets; their median energy efficiency is significantly lower than those of SUPERC and ULTRSC. 
The improvement in operational efficiency of coal power plants is strongly dependent on the real-time estimation of optimal values of the operating variables that affect thermal efficiency. The main steam pressure and the main steam flow rate are two critically controlled operating variables during the power generation operation of coal power plants \cite{cengel2001fundamentals}. 

The power generation from the power plant (Power MW) is increased or decreased within the generation capacity limits, and the turbine heat rate is calculated to monitor the thermal energy spent on producing an electrical unit. Using the compiled data set of global coal power plants, the contour plots are made with the main steam pressure (measured in MPa) and the main steam flow rate (measured in t / h) against power (in MW) and the turbine heat rate (kJ/kWh), as shown in Fig.~\ref{Fig:global coal}(b) and Fig.~\ref{Fig:global coal}(c) respectively. The plots depict the nonconvex and highly nonlinear function spaces for Power and turbine heat rate which are not only difficult to optimise from an engineering perspective but also require advanced analytics for handling the complexity bound with the operational excellence of coal power plants.         

With recent advances in information communication technologies (ICTs) for data collection and storage, large-scale industrial systems are equipped with large volumes of data banks that can help improve the operational efficiency and performance of industrial systems, which in turn can aid in reducing carbon emissions. This data-driven modelling is carried out effectively with machine learning (ML) and artificial intelligence (AI) algorithms that have demonstrated their effectiveness in improving the performance of industrial sectors such as energy \cite{ashraf2024data}\cite{ashraf2022artificial}\cite{eyring2024pushing}, manufacturing \cite{ishfaq2018investigation}\cite{ishfaq2023enhancing}, water \cite{salem2022predictive}\cite{zhang2023machine} and chemical process industries \cite{schweidtmann2024generative}\cite{zendehboudi2018applications}. With improved access to ICTs, big data and greater computing power, we can deploy data-driven decision-making for rapid industrial decarbonisation tasks in all sectors \cite{ashraf2024data, debnath2023harnessing}. Nonetheless, efforts have been limited in this area due to challenges in carrying out the domain-consistent analytics that can be implemented on large-scale industrial systems for industrial decarbonisation.

The intersection of data science and process engineering has shown that ML and AI can be used to estimate the set points for input operating variables, ensuring efficient energy use, reduced fuel consumption, minimal emissions discharge, and lower operating costs of engineering systems \cite{eyring2024pushing}\cite{kaack2022aligning}, which are critical to achieving the goals of industrial decarbonisation. In the case of thermal power plant optimisation, techniques like non-linear programming \cite{floudas1995nonlinear}, metaheuristic solvers including genetic algorithms \cite{deb2002fast}, particle swarm \cite{kennedy1995particle}, etc. are frequently reported in the literature. However, it is important to mention here that establishing the data-driven inequality constraints that accurately represent the operating characteristics of the industrial systems is a challenging task for the Machine Learning and OPTimization (MLOPT) framework and is a potential research gap in thermal power systems' optimisation. Another problem in defining the system-specific and data-driven constraint is that it is hard to work together with the industrial expert community, and there are not enough representations of domain knowledge \cite{larosa2024artificial} in the formulated optimisation problems. Thus, mathematically feasible but practically ineffective and domain-inconsistent solutions are estimated that cannot be implemented in the industrial environment, and are the major barrier to AI adoption in the industry \cite{van2022overcoming}. This creates a missed opportunity to embed data science-led decision-making in improving operating efficiency and achieving industrial decarbonisation at scale, both of which are essential for climate action. This research paper attempts to bridge the industry-academia liaison gap and overcome the research bottlenecks that contribute to producing actionable solutions for industrial decarbonisation of coal power plants.

A growing body of literature shows that MLOPT challenges can be solved by introducing human oversight and domain knowledge to better contextualise the data-driven results of ML analytics \cite{debnath2023harnessing}. The introduction of expert knowledge and interactive feedback with ML systems to improve their output is known as human-in-the-loop (HITL) ML \cite{mosqueira2023human,kumar2024applications, retzlaff2024human, debnath2025enabling}. We implement the HITL concept to solve the MLOPT problem for the operational excellence of a coal-based thermal power plant. We incorporate human guidance and experience into the formulation of the optimisation problem through an innovative HITL-led data-driven constraint, which effectively captures the variables' dependence. The data-driven constraint, embedded in the formulated MLOPT problem, guides the optimisation solver to estimate an effective solution, ensuring that the structural relationship of the operating variables stays within the human-defined threshold. This approach synergises with the HITL-ML approach, resulting in effective solution estimation and better decision-making for the considered application.

\section{HITL design for operational optimisation tasks}

Currently, HITL-ML has three main approaches, namely active learning, interactive machine learning, and machine teaching \cite{mosqueira2023human}. These subclasses of HITL-ML are based on who is in control of the learning process: model (active learning), shared (interactive machine learning) and humans (machine teaching). The three subclasses are learning-centric with respect to ML models that interact with humans. Another subclass of HITL-ML incorporates human feedback and insights to guide optimisation solvers for estimating the solution to the defined problem. This approach is frequently used in robotics \cite{slade2024human}, where humans update the constraints to guide the optimiser for estimating the solutions to the defined optimisation problems that are fed to the robot, and the performance of the robot-human interaction to achieve the defined objective is iteratively evaluated. Human guidance in establishing the scope of data-driven constraints that characterise the operating behaviour of industrial systems like coal power plants is critical to helping the optimiser estimate the effective solutions that are domain-compliant and implementable in the industrial environment.   

In the optimisation problem, we define a HITL-led data-driven constraint on the correlated features of the coal power plant's operation since the feature associations and their impact on the performance variables contribute to the domain knowledge and system understanding of the industrial professional. The data-driven constraints limit the deviation in the scaled values of the correlated features to a human-selected tolerance level as decided by the humans considering the operating characteristics of the industrial systems. We looked at an optimisation problem for a 660 MW superciritical thermal (coal-based) power plant that aimed to maximise thermal efficiency and minimise turbine heat rate. The optimisation problem with the embedded data-driven constraint is solved by the solver that must meet the constraints to estimate the solutions for the multi-objective optimisation problem. We also conduct a comparison between the solutions obtained for optimisation problems, both with and without the data-driven constraint, and evaluate the distribution similarity of the correlated features to determine their implementability in plant operations. 

We anticipate that introducing data-driven constraints into MLOPT analytics can introduce a new paradigm of efficient solution estimation for industrial analytics that can accelerate the domain-compliant AI adoption in industry, which has been historically slow. Our study contributes to this emerging field of machine intelligence for industrial decarbonisation through domain knowledge representation and human-centred analytics that facilitate the energy-efficient operation of industrial systems to reduce carbon emissions. 

\section{Data and Methods}

In this paper, we present the HITL-based design framework for utilising human expert domain knowledge in aligning the MLOPT to estimate the effective solutions that are implementable on the existing control layer of industrial systems. We have maximised thermal efficiency and minimised the turbine heat rate, two important performance parameters of thermal power plants through the HITL-MLOPT framework that supports the operation excellence of the 660 MW capacity coal power plant.  

\subsection*{Problem formulation and data collection}

Data-centric analytics are based on the data collected from a 660 MW supercritical thermal power plant. Two critically important plant-level performance parameters, that is, thermal efficiency (TE) and turbine heat rate (THR), measured in\% and kJ / kWh, respectively, are modelled on the operating variables of the power plant. The operating space for industrial power generation on a 660 MW power plant is significantly large; therefore, domain knowledge and literature review \cite{sharma2023data, haddad2021parameter, smrekar2010prediction, speight2013coal} are carried out to select the relevant operating variables for the two plant-level performance parameters. The operating variables selected to model TE and THR are as follows: coal flow rate (CFR – t/h), total air flow rate (TAF – t/h), main steam pressure (MSP – MPa), main steam temperature (MST – $ \degree \text{C} $), main steam flow rate (MSF – t/h), feed water temperature (FWT – $ \degree \text{C} $), reheat temperature (RHT – $ \degree \text{C} $), condenser vacuum (CV – kPa), and Power output of the plant (Power – MW). CFR and TAF are the operating parameters of the flue gas cycle that exchange the heat produced from fuel combustion in the air with the feed water for steam generation. The rest of the operating variables except Power are taken from feed water \& steam cycles that produce high enthalpy steam in the boiler. The high-pressure and high-temperature steam expands in the series of turbines (high-pressure, intermediate pressure and low-pressure) to drive the rotor of the steam turbine system, which is coupled with the generator for the electric power generation. The power produced corresponding to the steam conditions and the fuel consumption rate affects TE and THR; therefore, the effective set points of the operating variables should be estimated to ensure the efficient operation of the power plant.

A total of 1278 observations, averaged over one hour, and associated with operating variables are taken from the power plant for the development of the ML model. The operating ranges of the variables are significantly different; therefore, the feature scaling technique is applied to scale the data to [0, 1]. The mathematical expression for the feature scaling is given as:

\begin{equation}
X_i^{\text{scaled}} = \frac{X_i - (X_i)_{min}}{(X_i)_{max} - (X_i)_{min}} 
\end{equation}

Here, $X_i^{\text{scaled}}$ is the scaled variable, and ($X_i)_{min}$ and ($X_i)_{max}$ are the minimum and maximum values of $X_i$ respectively. The empiric cumulative distribution function (ECDF) provides the probability of observing the value of the variable. The correlated variables, on the same scale, have nearly the same ECDF profiles mapped against each other, which are in contrast to those of independent variables \cite{pernot2018probabilistic} and allow identifying the dependent-independent set of variables in the list of operating variables. The colinearity of the variables can be effectively computed using Pearson Correlation Coefficient ($PCC$) which measures the linear dependence between a pair of variables. The mathematical expression of $PCC$ is given as:

\begin{equation}
PCC_{xy} = \frac{\sum_{i=1}^N (x_i - \bar{x})(y_i - \bar{y})}{\sqrt{\sum_{i=1}^N (x_i - \bar{x})^2 \sum_{i=1}^N (y_i - \bar{y})^2}} 
\end{equation}

here, $x_i$ \& $y_i$ are the pair of variables having observations, $i = 1,2,3,\dots,N$. $\bar{x}_i$ and $\bar{y}_i$ are the mean values of $x_i$ and $y_i$ respectively. A strong positive correlation has a $PCC$ value of +1, while a $PCC$ value of -1 confirms the strong negative correlation between the pair of variables. Most of the operating variables of the thermal power plants are linearly dependent on each other, which is governed by the nature of industrial processes, and the association and relationships of variables build the domain expertise of the professionals in the power plant. Thus, $PCC$ information provides insight into the power generation operation that should be represented in MLOPT-driven industrial analytics to achieve the excellence of the power plant operation.

\subsection*{Model development and Variable significance analysis}

We use eXtreme Gradient Boosting (XGBoost) algorithm here\cite{shwartz2022tabular}. The algorithm takes advantage of the training of the learners to sequentially improve the predictive performance. A well-trained XGBoost model requires rigorous and extensive tuning of the hyperparameters. Eta, gamma, reg\_lambda, max\_depth, subsample, colsample\_bytree and n\_estimators are the critical hyperparameters of the XGBoost model. In this paper, the hyperparameters of the XGBoost model are tuned by the Tree-structured Parzen Estimator solver, available in the Hyperopt library in Python \cite{bergstra2013hyperopt}. A data split ratio of 0.8 and 0.2 is applied for data partition into training and testing sets for the development of XGBoost models for TE and THR. The performance evaluation of the model is made by the coefficient of determination ($R^2$) and root-mean-squared-error ($RMSE$) \cite{jamil2024machine, zhang2023machine} which are computed as:

\begin{equation}
R^2 = 1 - \frac{\sum_{i=1}^N (y_i - \hat{y}_i)^2}{\sum_{i=1}^N (y_i - \bar{y})^2}
\end{equation}
\begin{equation}
RMSE = \sqrt\frac{\sum_{i=1}^N (y_i - \bar{y_i})}{N}
\end{equation}

here, $y_i$, $\hat{y}_i$ are the true and model-predicted values, respectively; while, $\bar{y}_i$ represents the mean value of $y_i$. $R^2$ is taken as a measure of accuracy and ranges from 0 to 1. Whereas, $RMSE$ is the overall mean error for the data predicted by the model and should be minimised to ensure the good predictive performance of the model.

In addition to ML model training, it is imperative to quantify the uncertainty associated with model-based point predictions. Inductive conformal prediction is a valid prediction interval estimation technique \cite{angelopoulos2021gentle} and is implemented on top of the trained XGBoost models for TE and THR. The inductive conformal prediction technique provides validated prediction intervals with fairly strong theoretical guarantees \cite{dewolf2023valid}. A calibration dataset is obtained with similar data distribution as those of the training and testing datasets (data exchangeability) that computes the nonconformity score. The prediction intervals are estimated with the computed nonconformity score on $1 - \alpha$ confidence level for data coverage in the constructed prediction intervals. 

The black-box nature of the XGBoost model ascertains the feature importance that demonstrates the significance of the variable towards the point predictions made by the model. In this regard, Shapley Additive Explanations (SHAP) is an established technique for feature importance analysis for ML models in the explainable AI (XAI) contexts. The working mechanism of the SHAP technique assigns an importance value to each feature, considering cooperative games built on the features \cite{lundberg2017unified}. Finally, the feature importance list depicts the variable significance order toward the predictions based on the ML model and also presents the interpretability profile of the model \cite{hinterleitner2024enhancing}. 

\subsection*{Model-based optimisation of thermal power plant}

The function profile of the two performance parameters is nonlinear and non-convex; a multi-objective optimisation problem is formulated that attempts to maximise TE and minimise THR. Considering the non-linear nature of the objective function, a nonlinear programming framework is implemented to solve the optimisation problem with the bounds on the operating variables. The MLOPT-based optimisation problem without data-driven constraints is formulated as:

Objective function: 
\[
\min_{\mathbf{x}} f(\mathbf{x}) = -\text{Normalized } f_{\text{TE}}(\mathbf{x}) + \text{Normalized } f_{\text{THR}}(\mathbf{x})
\]
Subject to
\[
h(\mathbf{x}) = 0
\]
\[
\mathbf{x} = \{x_1, x_2, \dots, x_m\} \tag{5}
\]
\[
\mathbf{x} \in \mathbf{X} \subseteq \mathbf{R}^n
\]
\[
\mathbf{x}^L \leq \mathbf{x} \leq \mathbf{x}^U
\]

The objective function incorporates the normalised expression of TE and THR that produces values in the range of 0 to 1. A negative sign means that TE is to be maximised while THR is to be minimised with equal weight to the two performance parameters. $h(\mathbf{x})$ is the equality constraint that represents the XGBoost models trained for TE and THR. $\mathbf{x}$ is a set of operating variables, $x_1$, $x_2$,\dots, $x_m$ that are continuous in nature and are associated with performance parameters. The bounds on $\mathbf{x}$ are provided to limit the search space for solution estimation to $\mathbf{x}^L$ (lower bound) and $\mathbf{x}^U$ (upper bound).

\subsection*{HITL-MLOPT Design, Evaluation and Verification}

The optimisation problem defined on the ML model and goal-orientated solver may provide the solutions that violate the structural relationships between the variables and thus are ineffective to be implemented on top of the control layer of the industrial systems. Here, HITL appears to be a promising approach to introducing prior knowledge and process domain expertise into the optimisation problem that can guide the solver to estimate solutions, satisfying the embedded data-driven and system-specific constraints. 

The correlated features present in the operating variables have distinct operating structures and parametric dependence. The additional information on the correlated structures that exist in the input space should be incorporated into the optimisation problem to represent the operating characteristics of the system. The correlation information for the pair of correlated features is embedded as the constraint that tends to minimise the deviation, on the scaled value of the correlated variables, up to the tolerance as guided by humans considering their domain expertise. The additional data-driven constraint forces the solver to respect the variables' relationships and structures while estimating the solution for the MLOPT-based formulated optimisation problem. Setting the parametric tolerance value introduces the HITL approach in MLOPT analytics that can enhance the adaptability of AI for industrial analytics through human intervention and may improve the implementation of AI-driven solutions in the industrial environment. The improved formulation of the optimisation problem with the proposed data-driven constraint under the HITL-MLOPT approach for power plant operation is given as:

Objective function: 
\[
\min_{\mathbf{x}} f(\mathbf{x}) = -\text{Normalized } f_{\text{TE}}(\mathbf{x}) + \text{Normalized } f_{\text{THR}}(\mathbf{x})
\]
Subject to:
\[
h(\mathbf{x}) = 0
\]
\[
\mathbf{x} = \{x_1, x_2, \dots, x_m\} \tag{6}
\]
\[
\begin{aligned}
|x_i' - x_{i+1}'| &< \tau & \quad & \forall i=1,2,3,\dots,u, \quad x_i' \in \mathbf{x}
\end{aligned}
\]
\[
\mathbf{x} \in \mathbf{X} \subseteq \mathbf{R}^n
\]
\[
\mathbf{x}^L \leq \mathbf{x} \leq \mathbf{x}^U
\]
$|x_i' - x_{i+1}'|< \tau$ is the data-driven constraint embedded in optimisation problem (6) that tends to minimise the absolute deviation of the scaled value for the correlated features ($x_i'$) up to $\tau$ and $x_i'$ is the subset of $x_i$. The inequality constraint is expanded for all correlated features, and the optimisation problem, satisfying the embedded constraints, is solved to estimate the feasible and optimal solution for the power plant's operation. 

The two optimisation problems, i.e., (5) without the data-driven constraint and (6) with the data-driven HITL constraint for $\tau = 0.05, 0.10, 0.15, 0.20, 0.25,$ and $0.30$, are solved for the initial guesses by the COBYLA solver \cite{powell1994direct} in Python. The solutions obtained after solving the optimisation problems (5) and (6) are selected corresponding to quantile values of 0, 25, 50, 75, 95 and 100, and are systematically compared to investigate the impact of embedding the data-driven constraint on the quality of solution estimation. Finally, estimated solutions based on the HITL approach are also tested in the operation of the power plant to investigate their implementability on the existing control layer of the power plant. The successful implementation of a HITL-MLOPT-driven solution in the power plant operation promotes operational excellence that supports industrial decarbonisation.

\section*{Results}

\subsection*{Data-driven understanding of the coal power plant}

\begin{figure}[htp]
    \centering
    \includegraphics[width=1.0\linewidth]{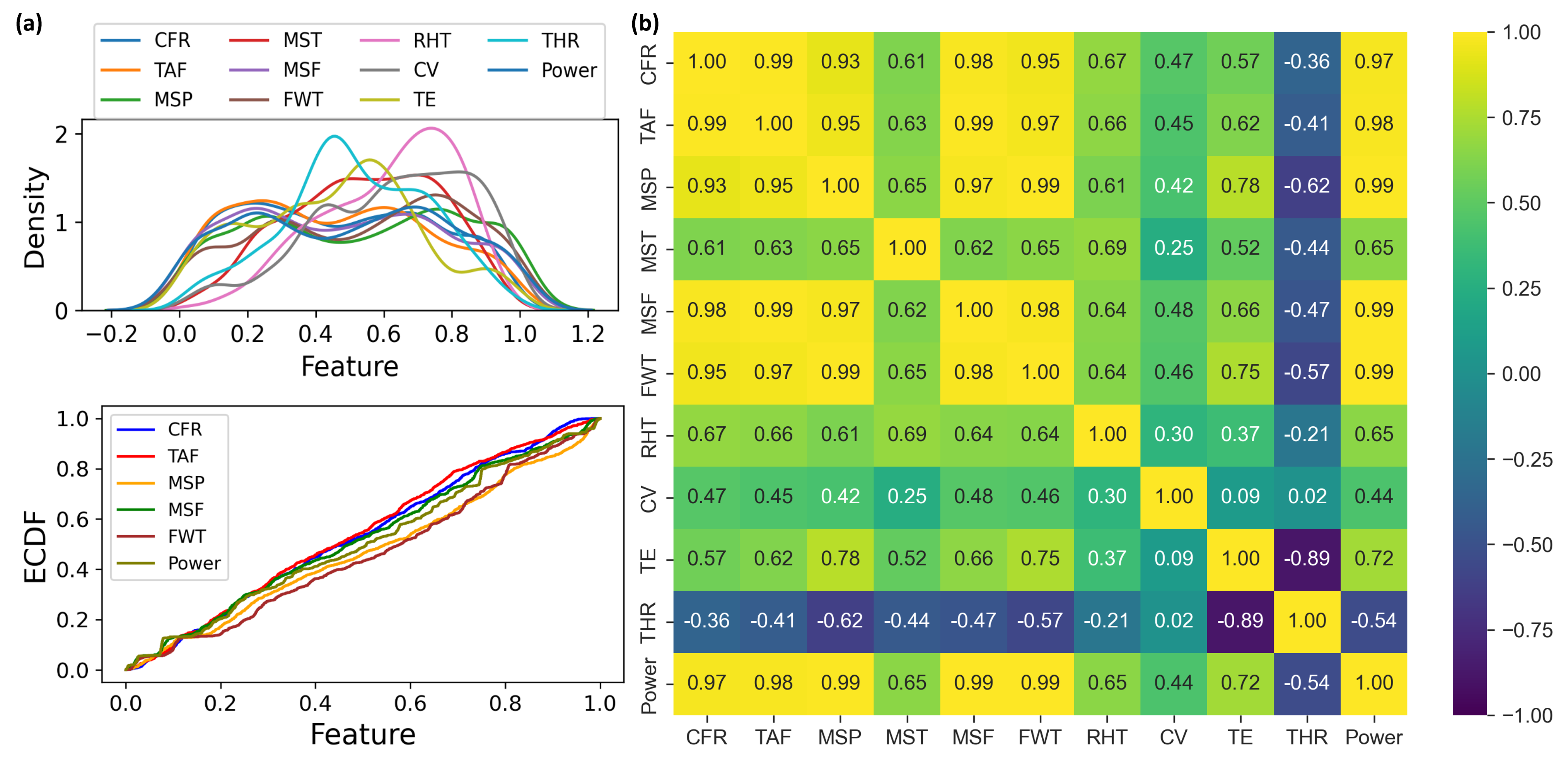}
    \caption{The operating behaviour of a 660 MW supercritical coal power plant. (a) The KDE curves for input-output features show that CFR, TAF, MSP, MSF, FWT and Power have nearly similar distribution profiles, meaning that these features are changed simultaneously during the plant operation. The ECDF profiles for the identified input-correlated features are close to each other, demonstrating the synchronised behaviour of the features. (b) The $PCC$ based heat map for the input-output variables quantifies the linear dependence between the pair of variables. The power generation operation is synchronised with the state of the variables in the control layer, resulting in the large number of correlated variables in the operating variables. The low $PCC$ value for the pair of variables shows that variables can be set at different set- points during the plant operation and are independent in nature.}
    \label{Fig:Data_Visual}
\end{figure}

Thermal power plants, especially coal-based plants, operate on a continuous heat exchange with the heating surfaces produced from fuel combustion to produce high-enthalpy steam that expands in the steam turbine system to generate power from the generator. The ramp-up and ramp-down operation of the coal power plant is driven by the amount of fuel supply in the boiler, and the operating variables associated with the coal flow rate (CFR - t/h) are adjusted accordingly. Thus, the temperatures, pressures, and flow rates of the state variables at different points of the power generation processes change during power generation. This understanding of the process is further confirmed by the data collected from the power plant that depict the typical operational characteristics of the plant operation in Fig.~\ref{Fig:Data_Visual}. 

The kernel density estimate (KDE) curves are plotted for the data associated with the variables and shown in Fig.~\ref{Fig:Data_Visual}(a). In the dynamic operation mode of the power plant, variables such as CFR, total air flow rate (TAF – t/h), main steam pressure (MSP – MPa), main steam flow rate (MSF – t/h), feed water temperature (FWT – $\degree \text{C} $), and generated power (Power - MW) have nearly the same KDE profiles that show the same operating trend and pattern about the change in these variables. The main steam temperature (MST – $\degree \text{C} $), reheat steam temperature (RHT – $\degree \text{C} $), and the condenser vacuum (CV - kPa) are also among the list of operating variables and their KDE curves are significantly different from those of CFR, TAF, MSP, MSF, FWT, and Power. The unique KDE curves of MST, RHT and CV demonstrate their independence for control purposes during plant operation, which are maintained according to the process dynamics and experience of human operators. 

We also constructed the empiric cumulative distribution function (ECDF) profiles on the scaled values of the variables to further investigate the dependence among similar KDE curves. It is observed that the ECDF profiles for CFR, TAF, MSP, MST, and Power are in close proximity to each other, which confirms the dependence and similar distributions among the features (see Fig.~\ref{Fig:Data_Visual}(a)). The way in which KDE curves and ECDF profiles are used to describe the plant's operation is in line with the power generation process \cite{sharma2023data}\cite{miller2010clean}, as the data show typical aspects of the industrial power generation process.

The Pearson correlation coefficient ($PCC$) calculated between the variables is illustrated as a heat map in Fig.~\ref{Fig:Data_Visual}(b). The features having $PCC$ $>$ 0.9 are taken as correlated pairs, and it is found that most of the input operating variables are linearly correlated, except MST, RHT, and CV. MSF, FWT, and Power have a moderate association with thermal efficiency (TE - \%) and turbine heat rate (THR – kJ/kWh) that is beneficial for the effective function approximation for the two performance parameters of the power plant. It should be noted here that the correlated variables, as included in the list of operating variables, are critically maintained in the operating windows and are particularly monitored during plant operation. Moreover, the association and dependence of variables, although for correlated variables, form the domain knowledge of industrial professionals about the process. Therefore, it is important to represent human expertise in domain-compliant analytics to improve their implementability in the industrial environment of the power plant. 

\subsection*{Model development and operation insights of coal power plant}

\begin{figure}[htp]
    \centering
    \includegraphics[width=1.0\linewidth]{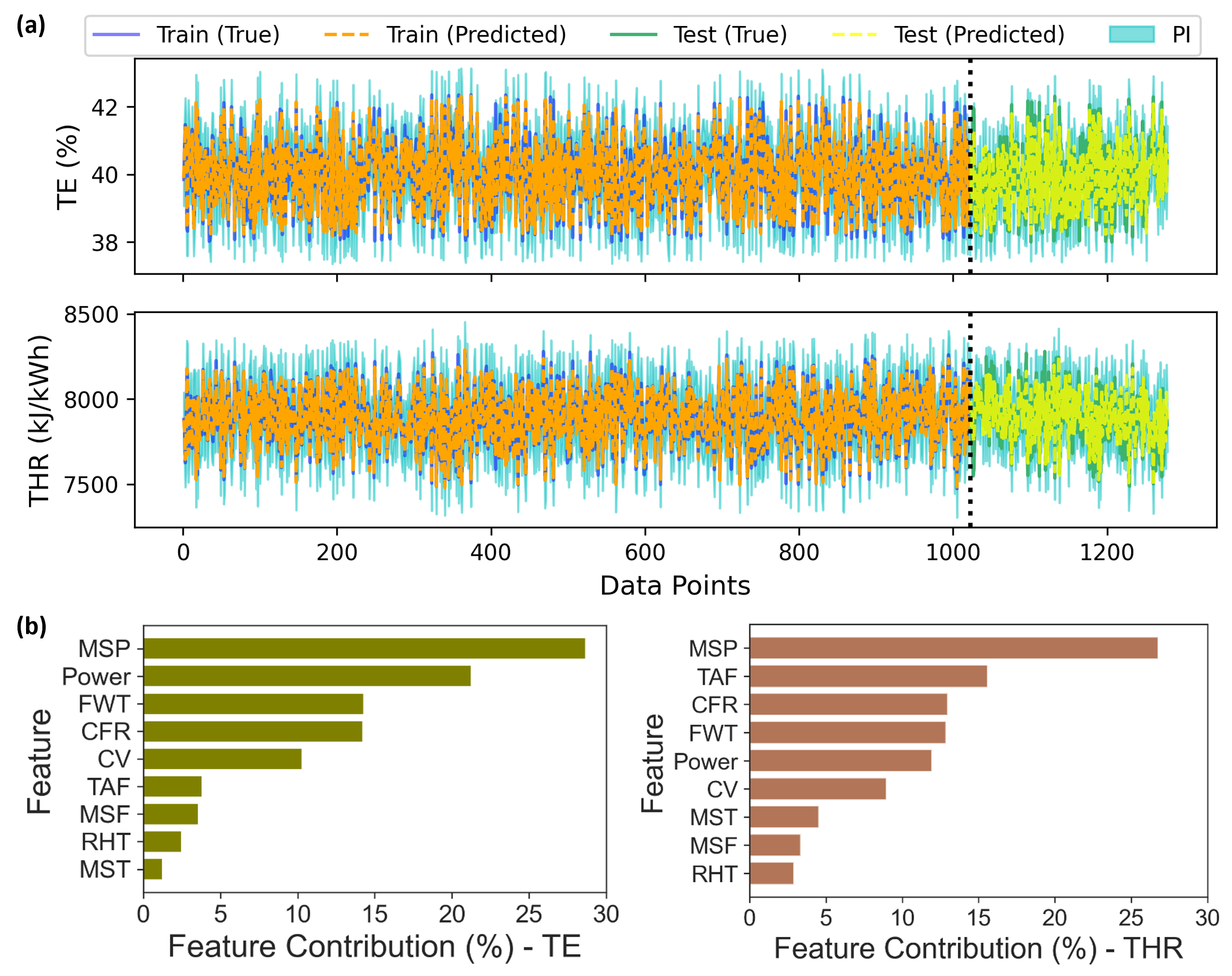}
    \caption{Data-driven modelling and insight into power generation operation. (a) XGBoost models are trained under rigorous hyperparameter tuning to predict TE and THR. The predictive performance of the models is evaluated in the training and testing datasets, and the prediction intervals are also constructed around the point-predictions of the models. (b) SHAP-based model interpretability analysis is carried out to compute the feature contribution towards the model-based predictions for TE and THR. A feature with a high feature contribution value is significant towards the model-based predictions. }
    \label{Fig:XGBoost}
\end{figure}

Data-driven modelling is performed with the XGBoost model to predict TE and THR of the coal power plant. The hyperparameters of the XGBoost model are rigorously tuned and the predictive accuracy of the trained models is evaluated by $R^2$ and $RMSE$ in the training and testing dataset. Fig.~\ref{Fig:XGBoost}(a) shows the true and XGBoost-based predicted values for TE and THR for training and testing datasets along with the prediction intervals drawn by the inductive conformal prediction technique on 95\% confidence level. 

The XGBoost model trained for TE predicts the training and testing dataset with $R^2$ of 0.96 and 0.88 and $RMSE$ of 0.2\% and 0.38\%, respectively. To go along with this, we find that the trained XGBoost model for THR makes predictions with a $R^2$ of 0.98 and $RMSE$ of 22.3 kJ/kWh on the training dataset and 69.5 kJ/kWh on the testing dataset. The prediction intervals with good data coverage are pretty tight on both the training and testing datasets for TE and THR. Overall, the trained XGBoost models for TE and THR depict good predictive performance, indicating that the models have learnt the data patterns and characteristics of the dynamic mode of plant operation. 

Understanding the model's interpretability is crucial for comprehending the prediction mechanism of ML models. Typically, a feature importance list investigates the interpretability of the model for regression-based problems, highlighting the significance of each feature in relation to model-based predictions. Fig.~\ref{Fig:XGBoost}(b) shows SHAP-based feature significance order along with the feature contribution measured in \% for TE and THR. MSP, Power and FWT are the first three significant features for thermal efficiency (TE), with feature contributions of 28.7\%, 21.3\%, and 14.3\%, respectively. However, MSP, TAF and CFR are the first three significant features that impact THR, with feature contributions of 26. 8\%, 15. 6\% and 13. 0\%, respectively. The increase in MSP increases the enthalpy and work potential of steam to drive the power generation process and significantly affects TE significantly. TE is characterised by the capacity discharge of the power plant and therefore depends on power generation. Moreover, FWT is maintained for the feed water entering the boiler, and it has a strong impact on the boiler operation and energy economy to produce steam under the specified conditions, and thus affects the TE of the power plant. Similarly, TAF and CFR are the flue gas parameters that produce the high-temperature flue gases that exchange heat with the working fluid. Inefficient heat exchange with heating surfaces during plant operation, as caused by numerous operating factors, drives the high thermal energy spent to ensure steam production corresponding to the state of the power generation, which can result in a higher THR of the power plant. The data-driven XGBoost model's interpretability analysis reveals that the feature importance order is well-aligned with the domain knowledge of plant operation. Additionally, the model's interpretability analysis shows how the XGBoost model makes predictions that a human plant operator can use their experience in the field to work with ML analytics and how the plant's operation responds to the dynamic variation in the operating variables.

\subsection*{HITL-led efficient solution estimation strategy}

The multiobjective optimisation problem that attempts to maximise TE and minimise THR and is subjected to bounds on the operating variables is defined in equation (5). The HITL-led data-driven constraint minimising the deviation of the scaled value for the two correlated features up to the defined tolerance is embedded in the optimisation problem (6). 

\begin{figure}[htp]
    \centering
    \includegraphics[width=1.0\linewidth]{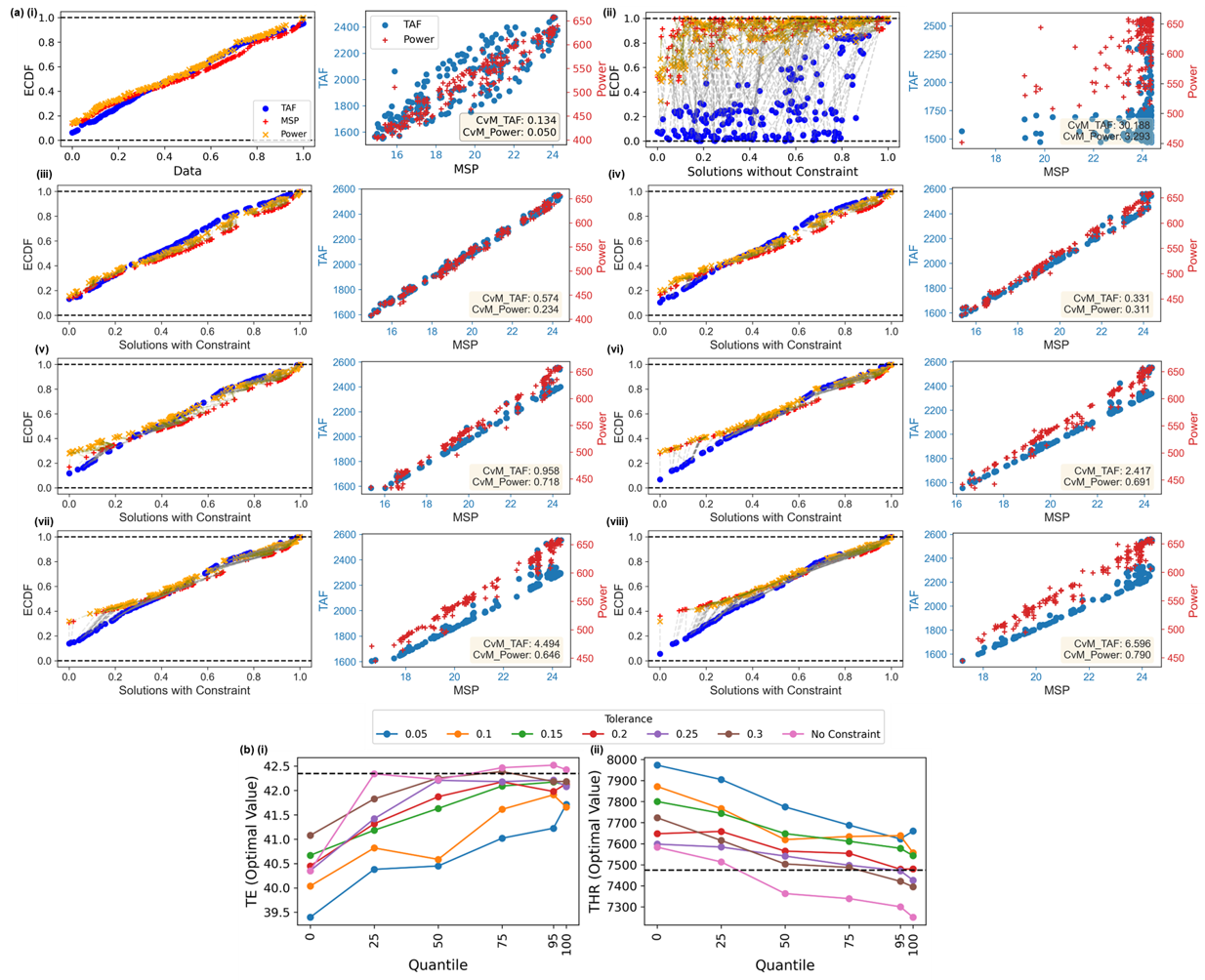}
    \caption{Comparison of multi-objective optimisation analysis that attempts to maximise TE and minimise THR for coal power plant by MLOPT framework. (a)(i) The ECDF profiles and scatter plot for two most significant correlated features for TE (MSP \& Power) and THR (MSP \& TAF) are plotted. The solution mapping for two correlated variables after solving the optimisation problem (5) is presented in (ii). The optimisation problem (6) is solved for a tolerance value of (iii) 0.05, (iv) 0.10, (v) 0.15, (vi) 0.20, (vii) 0.25, and (viii) 0.30. The optimal values of TE and THR estimated on the quantile value of the initial guesses are presented in (b) for (i) TE and (ii) THR.}
    \label{Fig:Opt}
\end{figure}

The two optimisation problems, formulated under the MLOPT framework, are solved for 219 initial guesses, which are historical data observations of the operation of the power plant. The ECDF profiles and scatter plot for the two most correlated characteristics, identified by SHAP analysis, for TE (MSP \& Power) and THR (MSP \& TAF) on the data of initial guesses are constructed and are shown on Fig.~\ref{Fig:Opt}(a)(i). The ECDF and scatter plots for MSP, Power and TAF are nearly organised around the diagonal, depicting the variable dependence that exists in the data. Furthermore, the Cramer-von Mises (CvM) metric computed between MSP \& TAF (CvM\_TAF) and between MSP \& Power (CvM\_Power) is 0.134 and 0.050, indicating that the distribution between the two pairs of features is nearly similar and serves as a benchmark to be compared with those of CvM distributions constructed for optimisation problems (5) and (6).

Fig.~\ref{Fig:Opt}(a)(ii) illustrates the ECDF profiles and scatter plot for MSP, TAF, and Power based optimal values obtained after solving the optimisation problem (5) corresponding to the initial guesses. The optimal solutions obtained for two of the most correlated features for the optimisation problem (5) have the ECDF profiles and the scatter plot, which are different from those of the data distribution profiles shown in Fig.~\ref{Fig:Opt}(a)(i). Furthermore, CvM\_TAF and CvM\_Power are computed to be 30.188 and 3.293, which confirms that the two distributions are potentially different from each other. Quantification of distribution dissimilarity of the correlated features with those of Fig.~\ref{Fig:Opt} (a) (i) confirms that the solutions obtained after solving the optimisation problem (5) are domain-inconsistent and thus,cannot be implemented in the existing control layer of the power plant; although these solutions are mathematically feasible but ineffective practically.

The optimisation problem (6) is solved with a tolerance value of 0.05 to 0.30 with a step size of 0.05. The tolerance can be set by the human operator, bringing human insights and process expertise in optimisation environment that enforces the solver to find domain-consistent solutions. The ECDF profiles and scatter plots, constructed on estimated optimal values, for two correlated variables of TE and THR are shown on Fig.~\ref{Fig:Opt}(a)(iii-viii), respectively. Note that the data distribution corresponding to ECDF profiles and scatter plots for the two correlated variables of TE and THR is similar up to the tolerance limit of 0.15, with CvM values less than 1.0. With the further increase in the tolerance allowance from 0.20 to 0.30, the ECDF and scatter plot distributions tend to deviate from each other, with CvM\_TAF reaching up to 6.596. However, CvM\_Power remains less than 1.0 for higher tolerance values, which means that the ECDF and scatter plot have nearly the same distribution as those of Fig.~\ref{Fig:Opt}(a)(i). The results shown in Fig.~\ref{Fig:Opt}(a)(iii-viii) highlight the need to investigate different tolerance values in the HITL-MLOPT analytics for the solution estimation that have nearly the same distribution for (i) the solver search space for the estimation of the domain-consistent optimal solutions and (ii) the implementation of the estimated solution in the existing control layer of the power plant operation. 

The impact of tolerances, as investigated using the HITL approach, on the estimated optimal values of the TE and THR is depicted in Fig.~\ref{Fig:Opt}(b)(i) and Fig.~\ref{Fig:Opt}(b)(ii), respectively. The dashed line on the TE and THR plots depicts the highest (42.35\%) and the lowest value (7575 kJ/kWh) of the performance parameters, respectively, that exist in the data obtained from the power plant. We take the solutions corresponding to the quantile values of 0, 25, 50, 75, 95, and 100 to systematically compare the optimal values of the TE and THR with different tolerance values.

We noted that the optimal values of TE and THR for lower tolerance values are relatively smaller than those of higher tolerance values, and the solver tries to reach the maximum and minimum values of TE and THR respectively, for tolerance greater than 0.10. It should be noted here that the optimisation solver is estimating a solution that maximises TE and minimises THR for the same operating values of the operating variables; thus, an optimal solution is achieved corresponding to minimisation of the objective function. It is interesting to note here that the optimal values estimated for optimisation problem (5) (no constraint) provides the solution with the bounds on the operating variables that are significantly far away from the dotted line. This is attributed to the parametric nature of the ML model, greedy nature of the solver to minimise the objective, and free exploration of the search space without considering feature distribution to be satisfied, resulting in estimating a solution that is mathematically feasible yet inefficient to be implemented on top of the existing control layer of the power plant.

\subsection*{Verification of HITL-MLOPT framework-driven solutions}

\begin{figure}[htp!]
    \centering
    \includegraphics[width=1.0\linewidth]{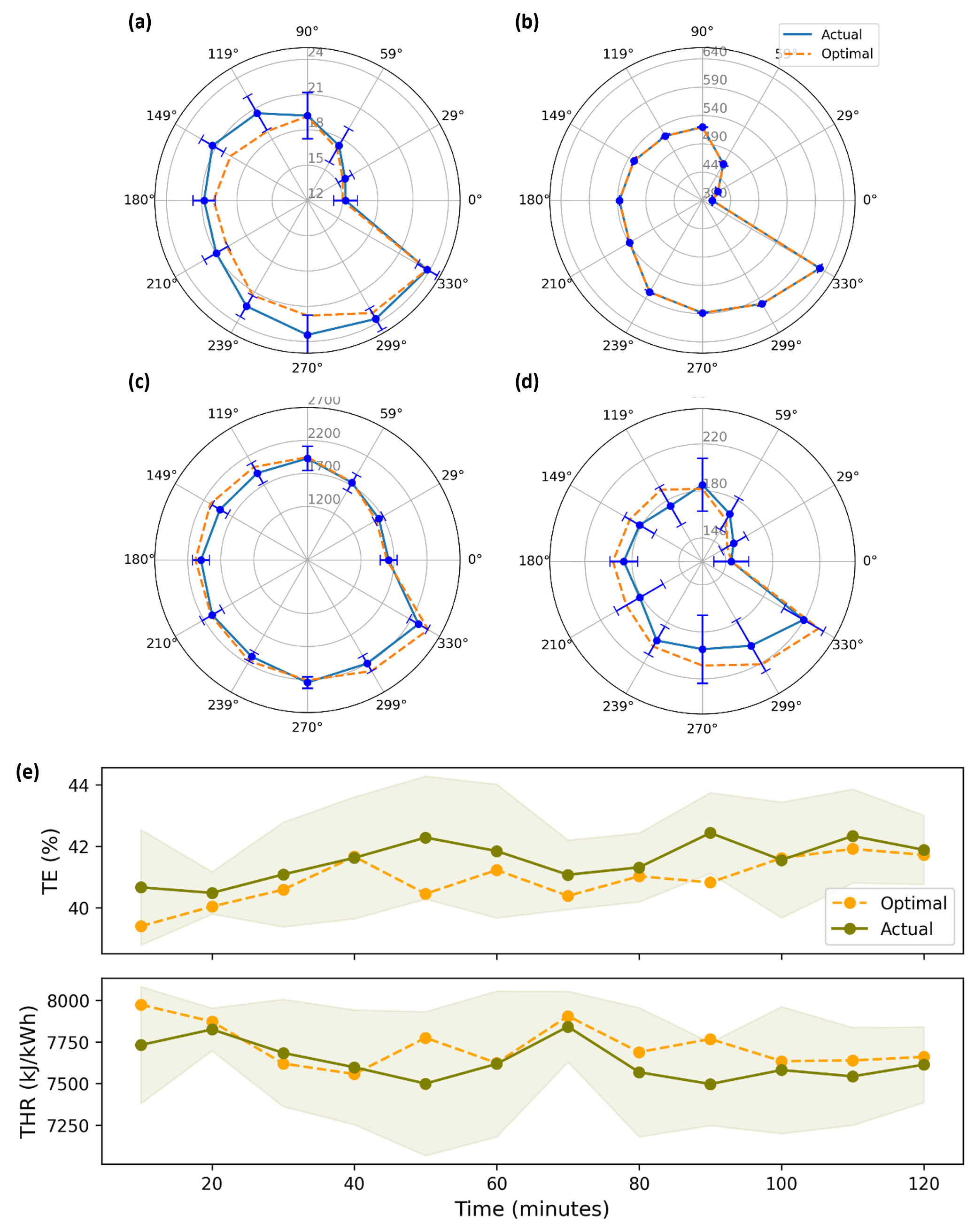}
    \caption{Verification of the HITL-MLOPT framework-driven estimated solutions on the power plant operation. The actual values of the operating vaiables (a) MSP (MPa), (b) Power (MW), (c) TAF (t/h) and (d) CFR (t/h) are plotted corresponding to the optimal values estimated from HITL-based MLOPT analytics. (e) The optimal values of TE and THR are  mapped against the actual values of TE and THR, maintained during the plant operation. The confidence intervals are drawn around the actual values with a 95\% confidence level.}
    \label{Fig:Testing}
\end{figure}

The estimated optimal operating values based on the HITL-MLOPT framework for the optimisation problem (6) are investigated in the plant operation. The operating conditions of the power plant are maintained close to the optimal estimated values of the operating variables. A ten-minute average (standard time interval for investigating the dynamic response of the system) for the operating values of the operating variables and performance parameters (TE and THR) is computed during the solution verification on the power plant's operation. This allows us to compare the actual values of performance parameters with the optimal values estimated through the HITL-MLOPT framework. The confidence interval at the 95\% CI is constructed to account for the variation in the actual values of the operating variables during plant operation. 

Fig.~\ref{Fig:Testing}(a–d) shows the plots for operating variables such as MSP, Power, TAF, and CFR, respectively, that significantly affect TE and THR. It should be noted here that the average values of the operating variables in 10 minutes (actual) are maintained close to the estimated optimal values of the operating variables (optimal). We observe that the computed confidence intervals for the actual values of Power and TAF are quite tight for the operating instances. The relatively larger prediction intervals in some cases for MSP and CFR are attributed to the logic structure implemented to control the power generation operation. CFR is maintained by six individual coal mills in the power plant, and each coal mill has the limit to supply coal to the furnace. As a new coal mill starts to reach the set value of CFR, the transient operation of the fuel injection disturbs the coal firing and combustion stability in the furnace, resulting in a variation in CFR that is depicted in the relatively larger prediction intervals for a few operating instances. The turbulent heat flux affects the steam generation process and the synchronised operation of the feed water pump, causing variation in pressure maintained by the feed water pump. It is also important to mention here that MSP variation is also affected by main steam temperature control using attemperation water flow rate, governing valve performance, and soot-blower efficiency. However, smooth thermodynamic steam conditions are maintained to ensure efficient power generation operation of the power plant. The complex network of hyperdimensional control space and the large-scale operation of the coal power plant make it difficult to maintain a smooth and safe power generation operation. However, during solution verification, we note that the optimal values of the operating variables fall within the constructed confidence intervals for MSP and CFR, which demonstrates that effective operation management is implemented by adjusting the operating variables near the optimal values, estimated by the HITL-MLOPT framework. 

During solution verification, the actual TE reaches the maximum value of 42.43\%, corresponding to the optimal value of 40.82\% as depicted on Fig.~\ref{Fig:Testing}(e). A slightly lower TE is also achieved during the solution verification phase, such as the actual TE of 41.55\% corresponding to the optimal TE value of 41.61\%. However, an average gain in TE is achieved by 0.64 percentage points corresponding to the testing instances during the solution verification phase. Similarly, the actual THR is maintained at the lowest value of 7499 kJ / kWh, corresponding to an optimal value of 7775 kJ/kWh, and the mean decrease in THR is measured to be 93 kJ/kWh for the testing instances.

\subsection*{Total Lifetime Reduction in CO$_2$ emissions in Global Coal Power Plants}

The optimal solutions estimated from the HITL-MLOPT analytics drive an average increase in TE by 0.64 percentage points for a 660 MW capacity thermal power plant operating on bituminous coal. We identified 56 supercritical, bituminous coal and 660 MW capacity power plants from the dataset of nearly 14200 global coal power plants \cite{Gem}. The design characteristics such as generation capacity, steam generation technology, type of fuel, and steam conditions of the identified thermal power plants are comparable to those of the power plant investigated for the HITL-MLOPT analysis in this paper. Therefore, the improvement at the plant level in the TE of the existing coal power plant can increase the TE of the identified coal power plants by 0.64 percentage points. The improved TE achieveable during the operation of thermal power plants is translated into a total reduction in CO$_2$ corresponding to the lifetime operation of the power plants calculated in \cite{ashraf2024driving}.  

\begin{figure}[htp!]
    \centering
    \includegraphics[width=1.0\linewidth]{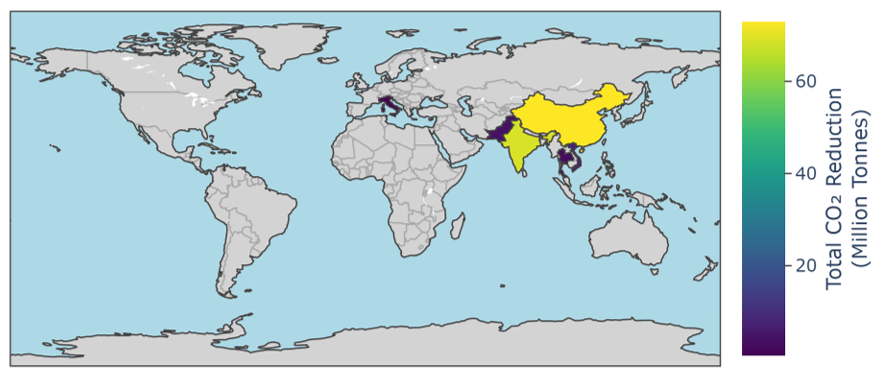}
    \caption{Total lifetime reduction in CO$_2$ from bituminous, supercritical, and 660 MW capacity coal power plants in the lifecycle operation. China takes the lead in cutting down total CO$_2$ followed by India,  Vietnam, Pakistan, Thailand, and Italy.}
    \label{Fig:global}
\end{figure}

The identified coal power plants are located in Pakistan, India, China, Italy, Thailand, and Vietnam, and the associated total lifetime reduction in CO$_2$ from the power plants is depicted in Fig.~\ref{Fig:global}. China leads the reduction in total CO$_2$ by up to 72.9 million tonnes from 26 coal power plants. India is the second biggest contributor, cutting down 68.8 million tonnes of CO$_2$ from 22 power plants. The coal power plants installed in Vietnam, Pakistan, Thailand, and Italy can reduce total CO$_2$ emissions, in million tonnes, up to 7.1, 4.1, 3.0 and 0.5 respectively. In doing so, 156.4 million tons of total CO$_2$ can be mitigated from the global coal power plants identified during their useful lifetime. 

Thus we show that a synergistic computational intelligence with human expertise and domain knowledge enhances the effectiveness and implementability of analytics that can be scaled up from the plant level to the global horizon to reduce CO$_2$ emissions to support industrial decarbonisation of thermal power plants.

\section*{Discussion and Conclusion}

The deployment of trained ML models into the optimisation problem, referred to as the MLOPT framework, can provide optimal solutions maximising the TE and minimising the THR of the thermal power plants. However, formulating the optimisation problem with the data-driven constraints that align with the human experience and capture the structural configuration of the variables is critical to estimating efficient solutions that are domain-consistent and are implementable on top of the existing control layer of industrial systems. 

We have proposed a HITL-led data-driven constraint that brings human insights and experience into the optimisation environment, and the defined constraint restricts the deviation in the scaled values of the correlated features up to a tolerance as specified by human operators. The data-driven constraint is embedded in the optimisation problem that the solver should satisfy while estimating the feasible solution. The optimisation problem without the proposed data-driven constraint is also solved with the same initial guesses, and the solutions obtained from two optimisation problems are compared. 

We observed that the HITL-led optimisation problem for tolerances of 0.05, 0.10, and 0.15 estimates the solutions with nearly the same distribution and retains the structural configuration of correlated features, i.e., MSP \& Power for TE and MSP \& TAF for THR. However, as the tolerance increases from 0.20 to 0.30, the distribution between the correlated features appears to deviate from the actual distribution similarity of the features. We also noted that solving the optimisation problem without the proposed data-driven constraint breaks the feature association and structural configuration of the correlated variables, resulting in estimating the solutions that are domain-inconsistent and cannot be implemented on top of the existing control layer of the power plant.

The tolerance is a human-guided parameter that can be set at different values for the estimation of the solution that complies with the limitation of the control layer of the power plant for the implementation of the solution driven by the HITL approach. Moreover, tolerance also allows the solver to explore the design space for solution estimation, which tends to minimize the objective function. 

The estimated optimal solutions obtained from the HITL-MLOPT framework are verified in the power plant power generation operation. The maximum TE of 42. 43\% and the minimum THR of 7499 kJ / kWh are achieved upon verification of HITL-MLOPT led  optimisation analysis, with a mean gain in TE of 0.64 percentage points and a mean reduction in THR of 93 kJ/kWh. In addition, 156.4 million tonnes of total CO$_2$ can be cut from a total of 56 globally installed coal power plants, operating on 660 MW capacity, supercritical technology and bituminous coal, from their lifetime operation. The data-driven constraint-led MLOPT framework introducing HITL can effectively estimate the efficient solutions that can be implemented in thermal power plants to improve TE and enhance the performance of thermal power systems. 

Through this study, we showed the potential to improve industrial system operational performance using data-driven and human-centric approaches. We ultimately show that domain expertise matters and that a human-in-the-loop design of ML analytics can be a way forward to enable more data science-led optimisation and operational excellence of industrial thermal power systems. In addition, our specific case study of a supercritical thermal power plant also demonstrates the implementation of data-science-led solutions on industrial systems, and the proposed HITL-MLOPT framework can be a step towards AI adoption in the industry that can support industrial decarbonisation and improve the techno-economic performance of industrial sectors. 

\section*{Limitations and future work}
The developed HITL-MLOPT framework is applied to enhance the operational excellence of a 660 MW capacity coal power plant, which is the main application limit of the analysis framework. In future work, the authors would extend the analytics to a broad range of power generation capacities of coal power plants. The authors would also investigate the effect of the data-driven constraints, built on information extraction from the data, on the quality of the optimized solution(s) and the subsequent verification of the real-time operation of power plants.    

\section*{Acknowledgement}
WMA acknowledges that this work was supported by The Alan Turing Institute’s Enrichment Scheme/Turing Studentship Scheme. WMA also acknowledges the funding (CMMS-PHD-2021-006) received from The Punjab Education Endowment Fund (PEEF) for his PhD at University College London, UK. RD thanks support from the Cambridge Humanities Research Grant, UKRI Responsible AI Grant, ai@cam AIDEAS grant, and Bill \& Melinda Gates Foundation [OPP1144]. 

\section*{Data availability statement}
The code and data used in this research are available at:

\url{https://github.com/Waqar9871/Data-Driven-Constraint-Optimisation.git}

\section*{Declaration of Competing Interests}
The authors declare no competing interests.

\section*{Author contribution}
W.M.A. - Conceptualization, Data Collection, Formal Analysis,  Writing - Original draft, V.D. - Supervision, Project Management. R.D. - Visualization, Writing - Review and Editing, Validation.

\bibliographystyle{unsrt}  
\bibliography{references}  



\end{document}